\documentclass[11pt]{article}
\usepackage{graphicx} % Required for inserting images
\usepackage{amsmath}
\usepackage{lipsum}
\usepackage{pdfcomment}

\usepackage[table,xcdraw]{xcolor} % Enable table colors and HTML color codes
\usepackage{subcaption}
\usepackage[utf8]{inputenc}
\usepackage{longtable}
\usepackage{graphicx}
\usepackage{float}
\usepackage{booktabs}
\usepackage{float} 
\usepackage{geometry}
\usepackage{lineno}
\usepackage{authblk}
\usepackage{tcolorbox}
\usepackage{rotating}
\usepackage{booktabs}
\usepackage{threeparttable}
\usepackage{geometry} % Optional: for full page layout control
\usepackage{lscape}
\usepackage{siunitx}  % For aligning numbers on the decimal point
\usepackage{adjustbox} % To adjust table width if needed
\usepackage{pgfplots}
\usepackage{caption}
\usepgfplotslibrary{groupplots}

\usepackage[numbers,sort&compress]{natbib}

\usepackage{hyperref}
\hypersetup{
  colorlinks=true,
  linkcolor=blue,       % 章节跳转链接（如目录、交叉引用）
  citecolor=blue,       % 文献引用
  urlcolor=blue         % 网页链接
}

\usepackage{cite}

\usepackage{array}
\usepackage{multirow}

\usepackage{amssymb}
\usepackage{pifont}

\usepackage[colorinlistoftodos]{todonotes}

\usepackage{wrapfig}

\usepackage{cite}

\usepackage{amsthm}
\usepackage{titlesec}
\usepackage{titletoc}
\titleclass{\subsubsubsection}{straight}[\subsection]
\newcounter{subsubsubsection}[subsubsection]
\renewcommand\thesubsubsubsection{\thesubsubsection.\arabic{subsubsubsection}}
\titleformat{\subsubsubsection}{\normalfont\normalsize\bfseries}{\thesubsubsubsection}{1em}{}
\titlespacing*{\subsubsubsection}{0pt}{3.25ex plus 1ex minus .2ex}{1.5ex plus .2ex}
\titleformat*{\section}{\fontsize{11}{12}\selectfont\bfseries}
\titleformat*{\subsection}{\fontsize{11}{12}\selectfont\bfseries}
\titleformat*{\subsubsection}{\fontsize{11}{12}\selectfont\bfseries}
\titleformat*{\paragraph}{\fontsize{11}{12}\selectfont\bfseries}
\titleformat*{\subparagraph}{\fontsize{11}{12}\selectfont\bfseries}

\titlecontents{subsubsubsection}
[8em] % adjust this value to fit your needs
{\small}
{\hyperlink{toc.\thecontentslabel}{\thecontentslabel.} }
{}
{\ \titlerule*[.5pc]{.}\contentspage}

\titlespacing*{\subsubsubsection}{0pt}{3.25ex plus 1ex minus .2ex}{1.5ex plus .2ex}

\usepackage{caption}
\usepackage{microtype}

\usepackage{enumitem}
\setlist{itemsep=0em}

\usepackage{fancyhdr} % For custom headers/footers
\usepackage{lastpage} % For total number of pages

\usepackage{listings}
\usepackage{xcolor}

\definecolor{codegreen}{rgb}{0,0.6,0}
\definecolor{codegray}{rgb}{0.5,0.5,0.5}
\definecolor{codepurple}{rgb}{0.58,0,0.82}
\definecolor{backcolour}{rgb}{0.95,0.95,0.92}

\lstdefinestyle{mystyle}{
    backgroundcolor=\color{backcolour},   
    commentstyle=\color{codegreen},
    keywordstyle=\color{magenta},
    numberstyle=\tiny\color{codegray},
    stringstyle=\color{codepurple},
    basicstyle=\ttfamily\footnotesize,
    breakatwhitespace=false,         
    breaklines=true,                 
    captionpos=b,                    
    keepspaces=true,                 
    numbers=left,                    
    numbersep=5pt,                  
    showspaces=false,                
    showstringspaces=false,
    showtabs=false,                  
    tabsize=2
}

\title{Deep Learning for Pavement Condition Evaluation Using Satellite Imagery}

\author[1]{Prathyush Kumar Reddy Lebaku}
\author[1]{Lu Gao, Ph.D.}
\author[2]{Pan Lu, Ph.D.}
\author[3]{Jingran Sun, Ph.D.}

\affil[1]{Department of Civil and Environmental Engineering, University of Houston \\
\texttt{\{plebaku, lgao5\}@central.uh.edu}}
\affil[2]{College of Business, North Dakota State University \\
\texttt{pan.lu@ndsu.edu}}
\affil[3]{Department of Civil and Environmental Engineering, University of South Florida \\
\texttt{jingransun@usf.edu}}

\date{}

\begin{document}

\maketitle

% Abstract (Do not insert blank lines, i.e. \\) 
\section*{Abstract}
Civil infrastructure systems covers large land areas and needs frequent inspections to maintain their public service capabilities. The conventional approaches of manual surveys or vehicle-based automated surveys to assess infrastructure conditions are often labor-intensive and time-consuming. For this reason, it is worthwhile to explore more cost-effective methods for monitoring and maintaining these infrastructures. Fortunately, recent advancements in satellite systems and image processing algorithms have opened up new possibilities. Numerous satellite systems have been employed to monitor infrastructure conditions and identify damages. Due to the improvement in ground sample distance (GSD), the level of detail that can be captured has significantly increased. Taking advantage of these technology advancement, this research investigated to evaluate pavement conditions using deep learning models for analyzing satellite images. We gathered over 3,000 satellite images of pavement sections, together with pavement evaluation ratings from TxDOT's PMIS database. The results of our study show an accuracy rate is exceeding 90\%. This research paves the way for a rapid and cost-effective approach to evaluating the pavement network in the future.

% Keywords
\noindent \textbf{Keywords}: Deep Learning; Satellite Image; Pavement Condition Evaluation; PMIS; Ensemble Learning

\section{Introduction}\label{introduction}

Civil infrastructure encompasses essential elements such as roads, bridges, buildings, and other physical structures and systems that are crucial for society's seamless functioning and progress \citep{gao2012network}. These components, which differ in their applications and objectives, play a vital role and must be engineered to last for an extended period. Therefore, the management of civil infrastructure is critical and is necessary to govern the planning, designing, construction, maintenance, and operation.

Managing infrastructure presents a range of challenges, notably in detecting failures and ensuring timely maintenance and rehabilitation \citep{kulkarni2003pavement,jahanbakhsh2016estimating, haas1978pavement, golabi1982statewide}. Insights from infrastructure management agencies highlight that early detection and preventative actions significantly enhance the longevity of infrastructure and cut down on maintenance expenses  \citep{madanat1993incorporating,gao2021detection,sundin2001artificial,camahan1987optimal,gao2012bayesian}. The adoption of large-scale monitoring tools is essential for efficient problem detection. Traditionally, damage assessments and evaluations required extensive fieldwork by inspection teams, a method both time-consuming and labor-intensive. However, advances in technology have introduced an array of efficient tools for damage detection, reducing the reliance on manual inspections and facilitating rapid identification of wear and tear. Notably, the analysis of satellite images has become a promising method for monitoring the condition and utilization of infrastructure. Recent progress in satellite technology, image analysis techniques, and computer vision has expanded the possibilities for infrastructure management.

To take the advantages of the recent technology advancement, this research explores the application of satellite image processing in pavement condition assessment using deep learning models. One contribution of this research is innovatively applying satellite imagery to evaluate the condition of pavements, offering a cost-effective and scalable alternative to traditional surveys conducted manually or with vehicles. To enhance the accuracy of pavement condition classification, the research utilizes various pre-trained deep learning models and the ensemble model, which combines predictions from multiple top-performing models, achieves an impressive accuracy of 93\% and an F1 score of 0.93. The practical implications of these findings highlight the potential of combining satellite imagery with deep learning for network-level pavement assessments. This approach could greatly reduce the need for resource-intensive on-site inspections.

The rest of this paper is organized as follows: the Literature Review section reviews the current models in pavement inspection and assessment; the Methodology section presents the proposed method; the Case Study section applies and evaluates the performance of the proposed method; and the Conclusions section summarizes the findings.

\section{Literature Review}

Pavement management systems play a crucial role in facilitating effective decision-making regarding pavement evaluation and maintenance \citep{gao2008robust,xu2022review,peng2025evaluating,yu2023pavement,wang2003decision,tighe2000incorporating,lebaku2025assessing}. Pavement inspection and assessment have also been made easier by the introduction of different machine-learning algorithms \citep{vemuri2020pavement}. For example, \citet{cha2017DeepLearningBased} introduced a deep convolutional neural network architecture using images taken with hand-held cameras for concrete crack detection, achieving an impressive 98.22\% accuracy. \citet{pan2018DetectionAsphaltPavement} employed UAV-captured pavement images and applied machine learning methods, attaining an accuracy of 98.78\% with support vector machines, 98.46\% with artificial neural networks, and 98.83\% with the random forest approach in identifying pavement deformations, such as cracks and potholes. \citet{fan2018AutomaticPavementCrack} demonstrated the superiority of deep learning over conventional machine learning and image processing methods, particularly through a CNN-based strategy for crack detection. \citet{chitale2020PotholeDetectionDimension} utilized the deep learning algorithm YOLO (You Only Look Once) to identify deformities in pavement and it was effective in detecting potholes and estimating their size. \citet{fan2020EnsembleDeepConvolutional} innovated further by proposing a CNN ensemble without pooling layers, based on probability fusion. Tested on two public crack databases, this method delivered remarkable results, surpassing traditional approaches with over 90\% precision and recall, and an F1 score exceeding 0.9, showcasing the potent capability of deep learning in enhancing pavement inspection and assessment methodologies. Similarly, \citet{ji2021ImageBasedRoad} exploited UAV-gathered images, employing a Deep Convolutional Neural Network (DCNN) to identify cracks, accurately measuring their width and location for a detailed risk assessment. \citet{maniat2021DeepLearningbasedVisual} explored the assessment of pavement quality utilizing Google Street View (GSV) images, analyzed through a Convolutional Neural Network (CNN). \citet{ahmadi2022IntegratedMachineLearning} utilized different machine learning techniques like a neural network, K-Nearest Neighbors, Support Vector Machine, Decision Tree, hybrid model, and Bagged trees  for crack detection in asphalt. \citet{sholevar2022MachineLearningTechniquesa} in their review paper present that machine and deep learning models significantly outperform traditional methods in pavement condition assessment, delivering faster, more accurate, and adaptable results. \citet{jiang2023PavementCrackMeasurement} introduce an innovative method for swift inspection of urban road cracks by harnessing aerial imagery in conjunction with deep learning techniques for 3D reconstruction and crack segmentation. The team employed a deep learning-supported segmentation network for efficient image processing, alongside an improved U-Net model to ensure precise crack detection and analysis. \citet{gagliardi2023DeepNeuralNetworks} employed Deep Neural Network (DNN) algorithms, including YOLO v7 and U-Net, to detect and assess the severity of pavement distress. These algorithms were specifically utilized for object detection and semantic segmentation, enhancing the accuracy of identifying pavement damages.

Over the past decade, the research community has explored utilizing satellite for pavement management. For instance, \citet{haider2010EffectFrequencyPavement} proposed a method using satellite image for pavement monitoring, envisioning improved highway maintenance and reduced reliance on traditional inspections. \citet{fagrhi2015SatelliteAssessmentMonitoring} reviewed various applications employing satellite imagery, including pavement management, where they analyzed historical data to detect deformations and assess deformation speed in highways, railways, and pavement irregularities. \citet{li2017EconomicFeasibilityStudy} conducted a cost-benefit analysis of pavement monitoring activities to study the financial aspects of this technology. Recently, the emergence of deep learning-based image processing methods has driven additional research into the efficacy of using satellite imagery to monitor pavement surface conditions. \citet{brewer2021PredictingRoadQuality} summarized the use of high-resolution satellite imagery with CNN to determine road quality, achieving an impressive accuracy rate of 80\%. \citet{bashar2022ExploringCapabilitiesOptical} highlighted the cost-effective and rapid nature of using satellite imagery to evaluate road conditions. They conducted assessments by analyzing spectral and texture information derived from satellite imagery, successfully identifying pavement distresses. \citet{jiang2021DevelopmentPavementEvaluation} proposes a pavement evaluation tool that used Deep Convolution Neural network to segment the highway images from Google Earth to detect longitudinal and transverse cracking. \citet{karimzadeh2022DeepLearningModel} explored the application of remote sensing and deep learning in evaluating road surface conditions. Utilizing Synthetic Aperture Radar (SAR) for quick data gathering and deep learning for detailed analysis, their research successfully forecasted the state of roads before and after the 2016 Kumamoto earthquake, reaching a precision level of 87\%.

Based on previous studies, this research aims to investigate the capability of using satellite imagery to monitor the condition of pavements, by linking satellite imagery data to pavement performance assessments provided by highway authorities. For this purpose, we accessed satellite imagery through the National Oceanic and Atmospheric Administration (NOAA). This imagery, sourced from the National Geodetic Survey's emergency database, offers a vast collection of high-resolution, geographically pinpointed images from significant storms and disasters over time. In a similar manner, the data on pavement condition assessments was acquired from the Texas Department of Transportation (TxDOT).

\section{Methodology}\label{methodology}
In this study, we investigated pre-trained models, utilizing various well-established architectures and also employed an ensemble learning strategy, combining predictions from a select few top-performing models to enhance overall performance. Figure 1 illustrates the workflow of our method, which consists of four main stages.

\begin{enumerate}
      \def\labelenumi{\arabic{enumi}.}
      \item
            Data preprocessing: We divided the dataset into training and test sets in an 8:2 ratio. For the training dataset, we use oversampling strategy to create a balanced dataset with an equal distribution across the five condition categories to ensure balanced coverage.
      \item
            Individual transfer learning models: To obtain suitable networks, we employed various pre-trained models such as VGG19, ResNet50, InceptionV3, DenseNet121, InceptionResNetV2, MobileNet, MobileNetV2, and EfficientNetB0. After evaluating their performance, we selected four top models as the base classifiers for the subsequent steps.
      \item
            Ensemble learning: We used a weighted voting method in this strategy to further enhance classification performance. The weight for each model was determined based on its accuracy.
      \item
            Performance evaluating: To measure the overall classification ability of the algorithm, we used various evaluation indicators such as accuracy, precision, recall, and F1-score.
\end{enumerate}

\begin{figure}[htbp]
      \centering
      \includegraphics[width=1\linewidth]{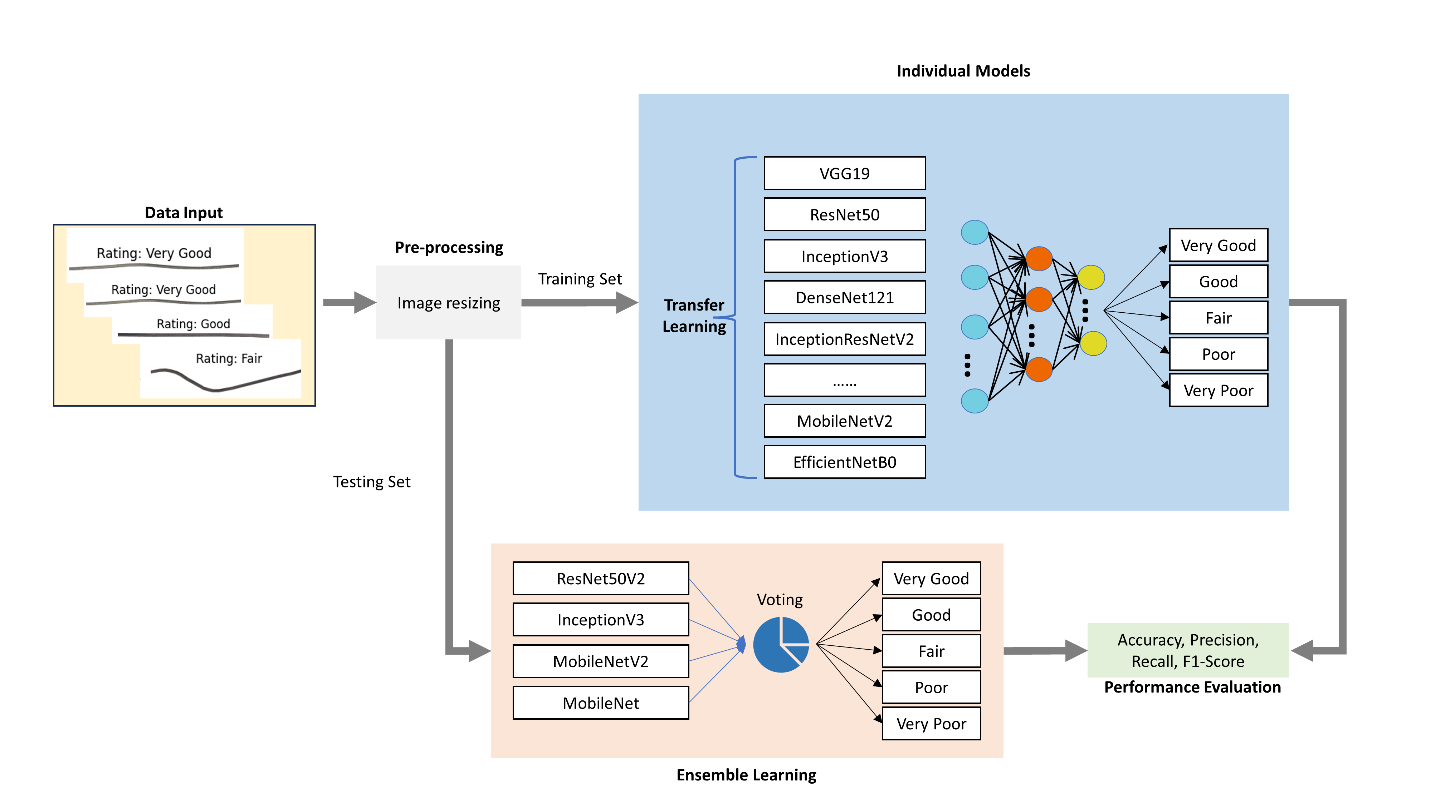}
      \caption{Proposed workflow}
      \label{fig:workflow}
\end{figure}

\subsection{Pre-Trained Models}\label{pre-trained-models}

Pre-trained models represent a powerful application of transfer learning within deep neural networks, particularly for image processing tasks. This technique involves utilizing a deep neural network trained on one dataset and transferring its learned weights to a new, related dataset. This approach expedites model training and enhances performance, especially when dealing with limited data.

The success of pre-trained models stems from the hierarchical nature of Convolutional Neural Networks (CNNs). Early layers capture low-level, generic features such as edges, textures, and shapes, which are widely applicable across different datasets. In contrast, later layers learn high-level features specific to the original dataset. Leveraging a pre-trained model allows these generic features to be transferred to a target dataset, enabling the model to adapt quickly to new tasks and conserving computational resources and time compared to training from scratch.

For our image processing project, we evaluated 16 popular pre-trained models, including ResNet50, VGG19, MobileNet, MobileNetV2, EfficientNetB0, DenseNet121, InceptionResNetV2, and InceptionV3. These models, extensively trained on large-scale datasets, have learned rich and transferable feature representations. Our performance evaluation involved running each model on our target dataset and assessing its accuracy, generalization capability, and ability to extract relevant features. The characteristics of these models are outlined in Table~\ref{table:models}.

\begin{table}[htbp]
      \caption{Summary of Pre-Trained Models}\label{table:models}
      \begin{tabular}{p{5cm}p{3cm}p{3cm}p{2cm}}
            \hline
            \textbf{Model}                                                        & \textbf{Parameters (Millions)} & \textbf{Year of Development} & \textbf{Depth (Layers)} \\ \hline
            VGG16 \citep{simonyan2015VeryDeepConvolutional}                       & 138                            & 2014                         & 16                      \\
            VGG19  \citep{simonyan2015VeryDeepConvolutional}                      & 143.7                          & 2014                         & 19                      \\
            InceptionV3 \citep{szegedy2016RethinkingInceptionArchitecture}        & 23.9                           & 2016                         & 159                     \\
            DenseNet121 \citep{huang2017DenselyConnectedConvolutional}            & 8.1                            & 2017                         & 121                     \\
            InceptionResNetV2 \citep{szegedy2017Inceptionv4InceptionResNetImpact} & 55.9                           & 2017                         & 572                     \\
            ResNet50 \citep{he2016DeepResidualLearning}                           & 25.6                           & 2016                         & 50                      \\
            ResNet50V2 \citep{he2016DeepResidualLearning}                         & 25.6                           & 2016                         & 50                      \\
            MobileNet \citep{howard2017MobileNetsEfficientConvolutional}          & 4.2                            & 2017                         & 88                      \\
            MobileNetV2 \citep{sandler2018MobileNetV2InvertedResiduals}           & 3.5                            & 2018                         & 88                      \\
            EfficientNetB0 \citep{tan2019EfficientNetRethinkingModel}             & 5.3                            & 2019                         & 69                      \\
            MobileNetV3Large \citep{howard2019SearchingMobileNetV3}               & 5.4                            & 2019                         & 88                      \\
            MobileNetV3Small \citep{howard2019SearchingMobileNetV3}               & 2.9                            & 2019                         & 88                      \\
            RegNetX \citep{radosavovic2020DesigningNetworkDesign}                 & Varies                         & 2020                         & Varies                  \\
            EfficientNetV2B0 \citep{tan2021EfficientNetV2SmallerModels}           & 7.1                            & 2021                         & 69                      \\
            Mobile\_ViT \citep{mehta2022MobileViTLightweightGeneralpurpose}       & 5.6                            & 2021                         & 20                      \\
            ConvNeXtBase \citep{liu2022ConvNet2020s}                              & 88                             & 2022                         & 53                      \\ \hline
      \end{tabular}
\end{table}

In this study, we utilized a pre-trained model for image classification training. We discarded the top classification layer of the pre-trained model, preserving the feature extraction layers responsible for capturing general patterns and image representations. Subsequently, we introduced a customized top classification layer consisting of global average pooling and dense layers. To adapt a pre-trained model for a different task, its architecture is employed to build a base model with existing weights. It is important to lock the pre-trained layers to avoid resetting their weights, as this would effectively mean starting the training process from the beginning. Additional trainable layers are introduced to leverage features from the original dataset for predictions on the new task. The pre-trained model, initially loaded without its final output layer, requires the addition of a new output layer, usually a dense layer with units matching the number of desired predictions. This setup enables the model to generate predictions on new datasets. To further enhance performance, fine-tuning can be employed, which involves unfreezing the base model and retraining it on the entire dataset with a very slow learning rate. Deep transfer learning for image classification operates on the principle that a model trained on a large and diverse dataset can effectively capture a broad representation of the visual world. By utilizing a model trained on such a dataset, the learned feature maps can be leveraged, eliminating the need to begin training from the ground up \citep{tan2018survey}.

The following table presents an overview of selected pre-trained models, detailing their parameters, development years, and depths. Each model possesses distinct attributes that make it ideal for various image processing tasks. The VGG16 and VGG19 models, developed in 2014, are known for their simplicity and depth, consisting of 16 and 19 layers respectively. They use small (3x3) convolution filters, which capture fine details in images, making them effective for image classification tasks despite their large parameter sizes (138 million for VGG16 and 143.7 million for VGG19). InceptionV3, with 23.9 million parameters and 159 layers, is part of the Inception series developed by Google in 2016. It introduced factorized convolutions and aggressive regularization, enhancing computational efficiency and decreasing the parameter count relative to previous models. DenseNet121, introduced in 2017, employs a dense connectivity pattern where each layer receives inputs from all preceding layers. With 8.1 million parameters and 121 layers, it mitigates the vanishing gradient problem and improves feature propagation while maintaining a smaller model size compared to other architectures. InceptionResNetV2 combines the Inception and ResNet architectures, featuring 55.9 million parameters and 572 layers. Developed in 2017, it merges the efficiency of Inception modules with the residual connections of ResNet, leading to improved training and convergence. ResNet50 and its updated version ResNet50V2, both with 25.6 million parameters and 50 layers, were introduced in 2016. Known for their use of residual blocks, these models mitigate the vanishing gradient problem, allowing for deeper networks and robust performance in various image classification tasks. MobileNet and MobileNetV2, developed in 2017 and 2018 respectively, are designed for mobile and embedded vision applications. They use depthwise separable convolutions to reduce parameters and computational cost, making them highly efficient for real-time applications. MobileNet has 4.2 million parameters and 88 layers, while MobileNetV2 has 3.5 million parameters and the same number of layers. EfficientNetB0, introduced in 2019 with 5.3 million parameters and 69 layers, is part of the EfficientNet family. It scales the network dimensions (depth, width, and resolution) uniformly using a compound scaling method, resulting in better performance and efficiency compared to traditional models. MobileNetV3Large and MobileNetV3Small, released in 2019, improve on the efficiency of the MobileNet series. MobileNetV3Large has 5.4 million parameters and 88 layers, while MobileNetV3Small has 2.9 million parameters and 88 layers. RegNetX models, introduced in 2020, have varying parameters and depths. They are designed to provide a simple, regular structure that can be easily adjusted to achieve optimal performance across a range of tasks, making them highly flexible. EfficientNetV2B0, part of the EfficientNetV2 series introduced in 2021, has 7.1 million parameters and 69 layers. It builds upon the original EfficientNet models with improved training speed and parameter efficiency, making it suitable for high-performance applications with limited computational resources. Mobile\_ViT, developed in 2021, combines the efficiency of MobileNet with the performance benefits of Vision Transformers (ViT). With 5.6 million parameters and 20 layers, it provides a lightweight model that performs well on a variety of image classification tasks. ConvNeXtBase, introduced in 2022, represents a modernized version of conventional CNN architectures. With 88 million parameters and 53 layers, it integrates several advancements from recent research to achieve cutting-edge performance on image classification benchmarks. These pre-trained models, each with their unique architectures and strengths, offer a range of options for enhancing image classification tasks through deep transfer learning. By leveraging these models, the process of training for new tasks becomes significantly more efficient and effective.

\subsection{Ensemble Learning Model}\label{ensemble-learning-model}

In this paper, we leveraged ensemble learning to improve the image classification performance of our dataset. We selected the top four performing models among those discussed earlier and applied an ensemble approach. Each model was equipped with custom top classification layers and trained independently on the training data, providing predictions for the test dataset. To create a combined prediction, we took advantage of the complementary strengths of these four models by averaging their individual predictions. This ensemble learning approach capitalized on the varied strengths of the individual models, leading to improved accuracy and robustness in image classification tailored to our specific task. By combining the strengths of these top models, we aimed to achieve better results than using a single model alone. We found that combining these pre-trained models through ensemble techniques significantly enhances classification accuracy by optimally extracting features. The ensemble model, designed for pavement evaluation, leverages fine-tuned transfer learning to precisely extract features, addressing the multi-class classification challenge. Furthermore, the predicted class label can be obtained by using Equation~\ref{equ:predictedClassLabel}.

\begin{equation}
      \label{equ:predictedClassLabel}
      \hat{y}=\underset{i\in\{1,\dots,c\}}{\operatorname{argmax}}\frac{\sum_{j=1}^k p_j(y=i|M_j,x)}k
\end{equation}

Where,

\(M_{j}\) = \(j\)-th model

\(k\) = number of the selected pre-trained models

\(p_{j}\left( y = i|M_{j},x \right)\ \)= prediction probability of a class value \(i\) in a data sample \(x\) using \(M_{j}\)

\subsection{Evaluation Indicators}\label{evaluation-indicators}

To evaluate the classification results, we used a confusion matrix and various indicators to assess the overall performance of the models. Accuracy (Equation~\ref{equ:accuracy}) represents the overall correctness of a model's predictions. Precision (Equation~\ref{equ:recall}) measures the proportion of true positive predictions of a class to all instances predicted as positive of the same class. Recall (Equation~\ref{equ:precision}) assesses the model's ability to capture all positive instances of a class and is calculated as the ratio of true positive predictions to all actual positive instances. The F1 score (Equation~\ref{equ:F1}), a harmonic mean of precision and recall, balances these metrics and is particularly useful when dealing with imbalanced class distributions or when both precision and recall are equally important.

\begin{equation}\label{equ:accuracy}
      \mathrm{Accuracy}=\frac{\mathrm{TP}+\mathrm{TN}}{\mathrm{TP}+\mathrm{TN}+\mathrm{FN}+\mathrm{FP}}
\end{equation}

\begin{equation}\label{equ:recall}
      \mathrm{Recall}=\frac{\mathrm{TP}}{\mathrm{TP}+\mathrm{FN}}
\end{equation}

\begin{equation}\label{equ:precision}
      \text{Precision}=\frac{\mathrm{TP}}{\mathrm{TP}+\mathrm{FP}}
\end{equation}

\begin{equation}\label{equ:F1}
      F1=\frac{2*\mathrm{Recall}*\mathrm{Precision}}{\mathrm{Recall}+\mathrm{precision}}
\end{equation}

\section{Case Study}\label{case-study}

\subsection{Data Collection}\label{data-collection}

\textbf{Pavement Image data collection}

\begin{figure}[htbp]
      \centering
      \includegraphics[width=1\linewidth]{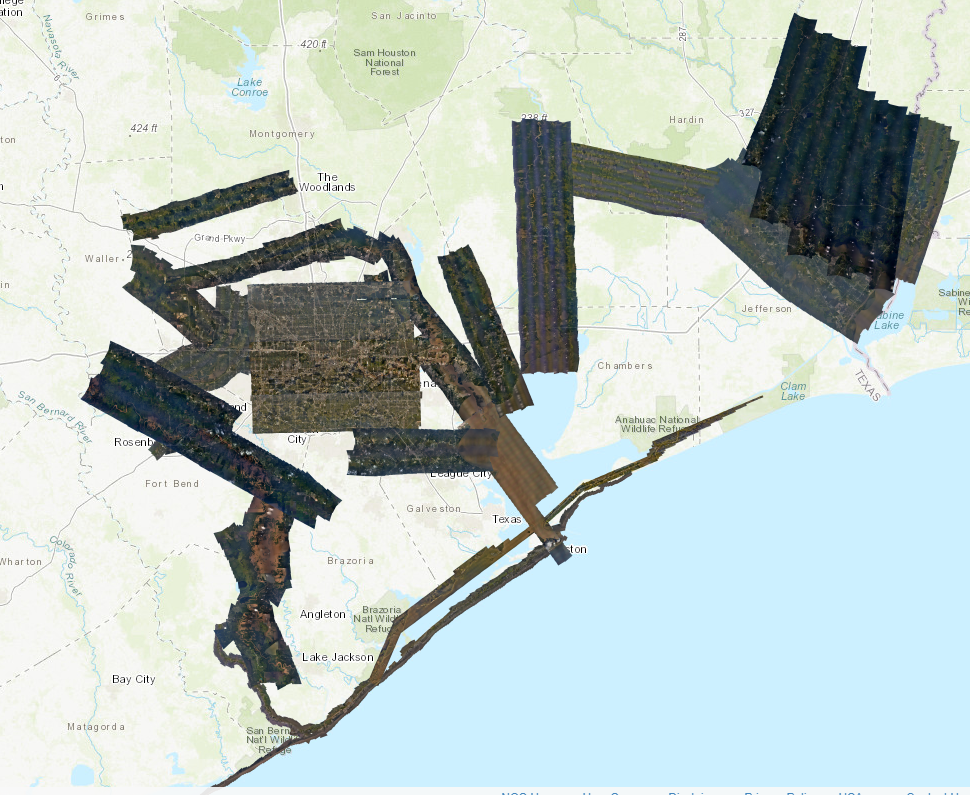}
      \caption{Satellite images coverage}
      \label{fig:coverage}
\end{figure}

In this case study, we utilized satellite imagery obtained from the National Oceanic and Atmospheric Administration (NOAA) with a resolution of 50 cm per pixel~\citep{noaa2023HurricaneHarveyImagery}. The images were specifically captured in September 2017 from the Houston metropolitan area. These images were originally collected by NOAA to document Hurricane Harvey, which struck Texas as a Category 4 hurricane on August 25, 2017. Figure~\ref{fig:coverage} offers a comprehensive view of the areas encompassed by the satellite images.

Commercial high-resolution satellite imagery can be quite expensive, with costs for 30-50 cm resolution images typically ranging from \$10 to \$20 per square kilometer. For this case study, which focuses on the road network within the Houston metropolitan area, covering approximately 26,000 square kilometers, the use of commercial satellite imagery would incur significant costs. However, following Hurricane Harvey, NOAA released high-resolution satellite images to the public at no charge. This invaluable access allowed us to leverage unique data for our research, an opportunity that would have been prohibitively expensive without access to these free images.

The extraction of road segments from the satellite images involved several steps. Initially, we acquired the coordinates of the centerline of the pavement network from the Texas Department of Transportation (TxDOT). By using interpolation methods, we obtained the centerline coordinates of the entire pavement network with an accuracy to the level of feet. This allowed us to match the coordinates with the reference markers of the pavement network accurately. For any pavement section defined in TxDOT's pavement management system, we could determine the exact coordinates of that section.

Subsequently, for each pavement section in the TxDOT pavement management dataset, typically 0.5 miles in length, a polygon was generated by extending 12 feet on each side of the centerline, effectively outlining the lane's coverage area. The satellite image provided by NOAA is georeferenced, meaning that each pixel of the image corresponds to a specific coordinate. Utilizing the polygons as guides, we precisely cropped out the corresponding portions of the georeferenced satellite images that represented the pavement sections. This process ensured that only the relevant parts of the satellite imagery were retained for further analysis. Using this approach, we obtained images of more than 3,000 pavement sections. The next step was to match these pavement sections with the pavement condition evaluation.

\textbf{Pavement condition data collection}

\begin{figure}[htbp]
      \centering
      \includegraphics[width=0.5\linewidth]{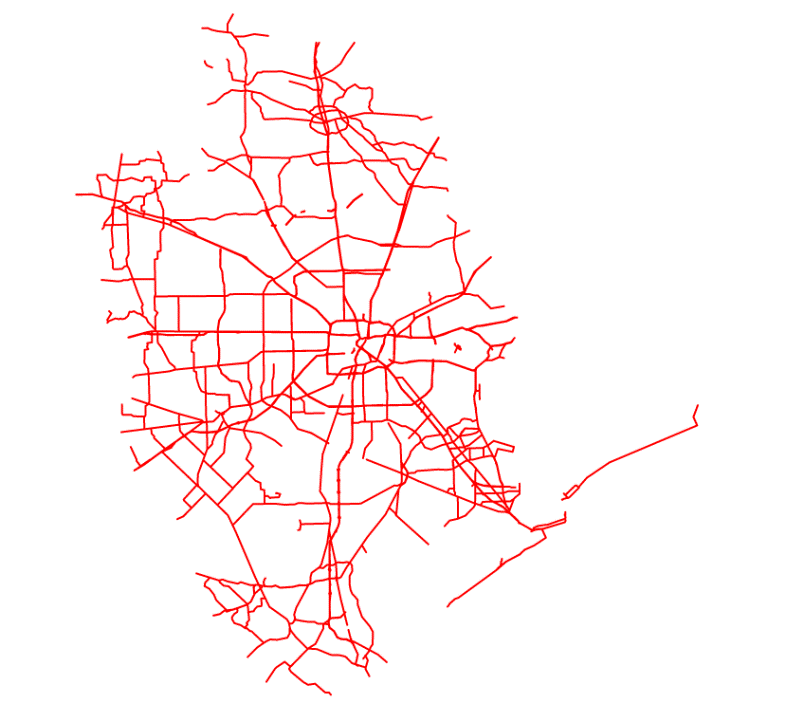}
      \caption{Pavement network data used in this case study}
      \label{fig:pavement}
\end{figure}

Pavement condition data for this study were obtained from the Texas Department of Transportation (TxDOT). Figure~\ref{fig:pavement} shows the road network managed by TxDOT's Houston district, and the data were specifically sourced from the Pavement Management Information System (PMIS) database. For this research, we utilized PMIS data corresponding to the same year from which satellite images were obtained from NOAA. Each pavement section in the PMIS database is uniquely identified by the columns ROUTE NAME, OFFSET FROM, and OFFSET TO, with most sections being approximately 0.5 miles long, though the exact length can vary from section to section. The starting and ending points of each section are fixed and remain consistent year over year. The coordinates of these points are available, allowing us to precisely locate each pavement section within the georeferenced satellite images, where every pixel is tagged with specific coordinates. This geospatial data enables us to accurately crop the pavement sections from the satellite images using their coordinates and then match each section's image with its corresponding Condition Score provided by TxDOT's annual evaluations.

We used the Condition Score in PMIS as the pavement condition indicator in this research. In the PMIS, there are three primary indicators reflect the general condition of the road pavement: the Distress Score, Ride Score, and Condition Score. The Distress Score, ranging from 1 (most severe distress) to 100 (minimal distress), evaluates visible surface deterioration \citep{gao2023deep}. This score combines various distress types to assess the impacts of cracking and rutting comprehensively. TxDOT calculates the overall distress score using a multiplicative utility analysis, which converts each specific distress type into a utility value (ranging from 0 to 1) via a formula:

\begin{equation}
      U_i=1-\alpha e^{-(\rho/L_i)^\beta}
\end{equation}

where,

\(U_{i} =\) Utility value of distress type \(i\)

\(L_{i} =\) Length of the distress type \(i\)

\(\alpha,\ \rho\ and\ \beta =\) Coefficients controlling the shape of
the curve and value of \(U_{i}\)

The overall distress score is calculated by multiplying 100 by the utility values associated with each type of distress relevant to the pavement category of the data collection segment, and is included in the Condition Score.

The Ride Score quantifies the ride quality of a pavement section on a scale from 0.1 (roughest) to 5.0 (smoothest). It is determined by calculating the length-weighted average of raw serviceability index (SI) values collected from the area, providing a comprehensive evaluation of ride comfort. 

The Condition Score offers an integrated evaluation of a pavement's state, combining the distress score and ride score into a single value ranging from 1 (poorest condition) to 100 (best condition), presented in Eq.~\ref{equ_CS}. This metric captures the general public's perception of the pavement's overall condition. TxDOT classifies pavements into five condition states based on their Condition Score, as shown in Table~\ref{table:CSclasses}, and these states are used for labeling each satellite image. 

\begin{equation}\label{equ_CS}
    CS = U_{ride} \times 100 \times \prod_i^{n} U_i
\end{equation}

where,

\(U_{i} =\) Utility value of distress type \(i\)

\(U_{ride} =\) Utility value of ride score

\begin{table}[htbp]
      \centering
      \caption{PMIS Condition Score Classes}
      \label{table:CSclasses}
      \begin{tabular}{cc}
            \hline
            Condition Score & Description \\ \hline
            90-100          & Very Good   \\
            70-89           & Good        \\
            50-69           & Fair        \\
            35-49           & Poor        \\
            1-34            & Very Poor   \\ \hline
      \end{tabular}
\end{table}

\begin{figure}[htbp]
      \centering
      \includegraphics[width=1\linewidth]{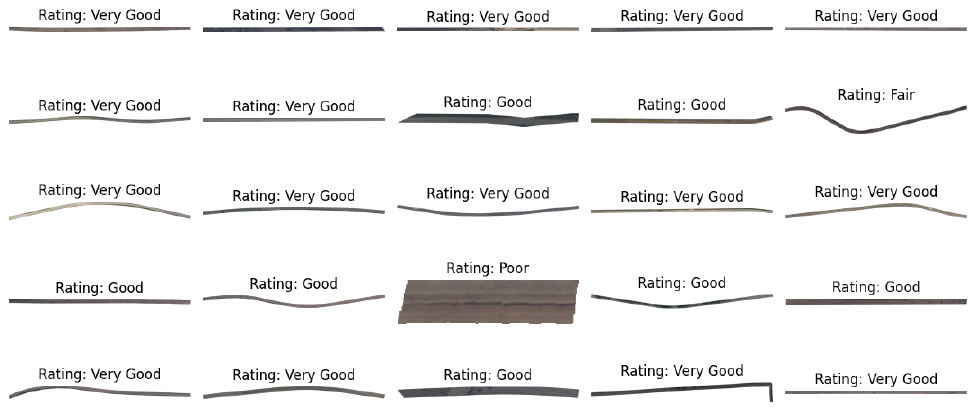}
      \caption{Sample cropped satellite images}
      \label{fig:images}
\end{figure}

\begin{figure}[htbp]
    \centering
    \begin{subfigure}[b]{0.9\textwidth}
        \centering
        \includegraphics[width=\textwidth]{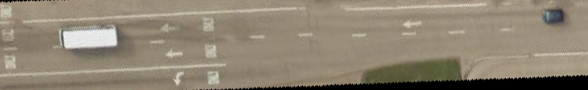}
        \caption{Very Good}
        \label{fig:sub1}
    \end{subfigure}
    
    \vspace{0.5cm} % Adjusts the space between subfigures

    \begin{subfigure}[b]{0.9\textwidth}
        \centering
        \includegraphics[width=\textwidth]{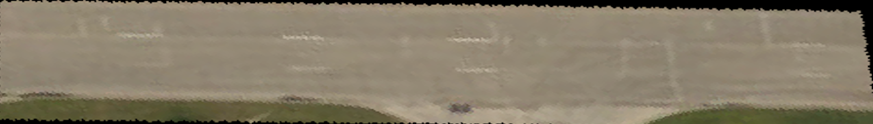}
        \caption{Good}
        \label{fig:sub2}
    \end{subfigure}
    
    \vspace{0.5cm} % Adjusts the space between subfigures

    \begin{subfigure}[b]{0.9\textwidth}
        \centering
        \includegraphics[width=\textwidth]{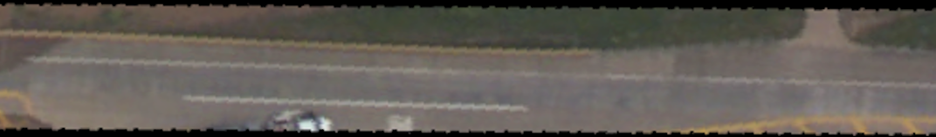}
        \caption{Fair}
        \label{fig:sub3}
    \end{subfigure}

    \vspace{0.5cm} % Adjusts the space between subfigures

    \begin{subfigure}[b]{0.9\textwidth}
        \centering
        \includegraphics[width=\textwidth]{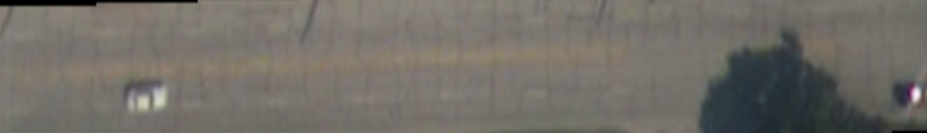}
        \caption{Poor}
        \label{fig:sub4}
    \end{subfigure}
    
    \vspace{0.5cm} % Adjusts the space between subfigures

    \begin{subfigure}[b]{0.9\textwidth}
        \centering
        \includegraphics[width=\textwidth]{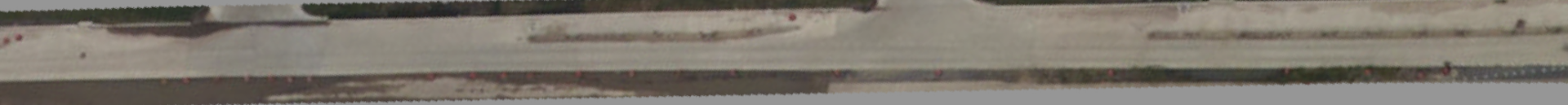}
        \caption{Very Poor}
        \label{fig:sub5}
    \end{subfigure}

    \caption{Close-up View of Pavement Segments Across Condition Categories}
    \label{fig:images2}
\end{figure}

Figure~\ref{fig:images} presents sample cropped satellite images from the datasets utilized in this research. Each image corresponds to a pavement management unit within the TxDOT Pavement Management System. These units are assigned a condition score based on annual inspections. As illustrated in Figure~\ref{fig:images}, the length and shape of each pavement segment vary. Figure~\ref{fig:images2} provides a closer view of segments from five different pavement sections categorized as very good, good, fair, poor, and very poor. As evident in Figure~\ref{fig:images2}, the image processing algorithm effectively learns from these images and identifies patterns in the pixel changes, thereby establishing a correlation with the condition evaluation results.

\subsection{Result and Analysis}\label{result-and-analysis}

\begin{figure}[htbp]
      \centering
      \includegraphics[width=1\linewidth]{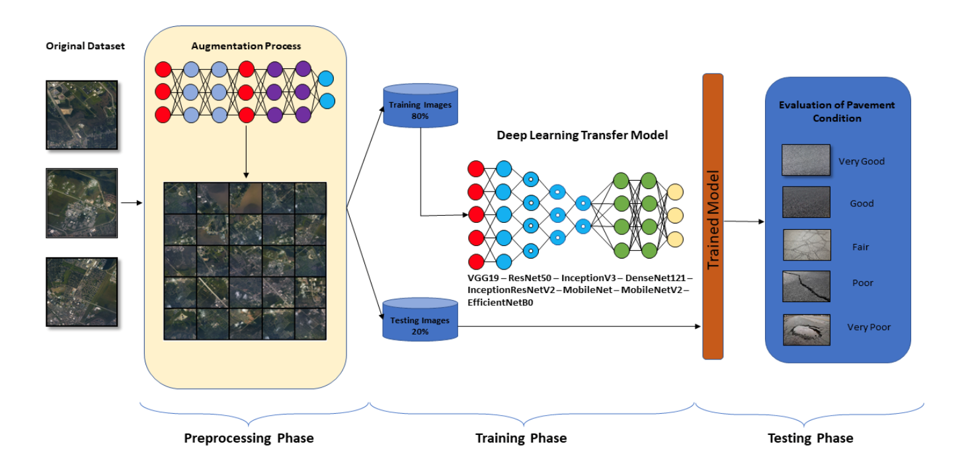}
      \caption{Satellite Image-Based Pavement Evaluation}
      \label{fig:flowchart2}
\end{figure}

In this study, we employed TensorFlow and Keras to construct and train our models, incorporating data augmentation techniques to enhance the training process. Data augmentation was used to artificially increase the diversity of our dataset by applying random transformations such as rotations, flips, zooms, shifts, and brightness adjustments to the original images. This approach helps prevent overfitting by allowing the models to generalize better from a more varied set of training examples. Figure \ref{fig:flowchart2} shows the process of evaluating pavement conditions using satellite images through a deep learning model, organized into three phases: Preprocessing, Training, and Testing. In the Preprocessing Phase, a dataset of satellite images, containing various views of roads and pavements, undergoes augmentation to create diverse variations. The augmented dataset is then split into 80\% for training and 20\% for testing. During the Training Phase, the training images are fed into a deep learning transfer model that utilizes pre-trained architectures, which are fine-tuned to learn specific features related to pavement conditions. In the Testing Phase, the trained model is evaluated using the remaining images to assess its accuracy in predicting pavement conditions, categorizing them into levels ranging from Very Good to Very Poor. 

The training process was carried out for 50 epochs on all models, using the default learning rate of the Adam optimizer. Each training iteration involved processing batches of 32 samples. Table~\ref{table:performance} presents a comprehensive comparison of the models' performance, evaluating both accuracy and F1 score. Notably, MobileNet and ResNet50V2 achieved the highest accuracies, with both models attaining accuracy over 89\% and F1 scores of 0.91 and 0.90, respectively. InceptionV3, DenseNet121, and MobileNetV2 also performed exceptionally well, each achieving accuracies around 87\% and F1 scores between 0.87 and 0.88. These results indicate that these models are highly effective for image classification tasks. In contrast, the VGG19 model, despite being a well-known architecture, achieved a relatively lower accuracy of 72\% and an F1 score of 0.73. This lower performance is likely due to the model's large number of parameters, which may result in overfitting. Similarly, the VGG16 model showed slightly better performance with an accuracy of 75\% and an F1 score of 0.74. Models like EfficientNetB0 and RegNetX demonstrated significantly lower performance, with accuracies around 20\% and 18\%, respectively, and corresponding F1 scores. These results suggest that these models may not be well-suited for the specific dataset or task at hand without further fine-tuning. Other models, such as MobileNetV3Large and MobileNetV3Small, showed moderate performance with accuracies of 58\% and 31\%, respectively. The ConvNeXtBase model also performed relatively well, achieving an accuracy of 78\% and an F1 score of 0.76.

% \textbf{Table 3. Performance of different models}
\begin{table}[htbp]
      \centering
      \caption{Performance of different models}\label{table:performance}
      \begin{tabular}{p{7cm}cc}
            \hline
            \textbf{Model}                                                        & \textbf{Accuracy} & \textbf{F1 Score} \\ \hline
            VGG16 \citep{simonyan2015VeryDeepConvolutional}                       & 0.75              & 0.74              \\
            VGG19 \citep{simonyan2015VeryDeepConvolutional}                       & 0.72              & 0.73              \\
            InceptionV3 \citep{szegedy2016RethinkingInceptionArchitecture}        & 0.87              & 0.88              \\
            DenseNet121 \citep{huang2017DenselyConnectedConvolutional}            & 0.86              & 0.88              \\
            InceptionResNetV2 \citep{szegedy2017Inceptionv4InceptionResNetImpact} & 0.86              & 0.87              \\
            ResNet50 \citep{he2016DeepResidualLearning}                           & 0.39              & 0.32              \\
            ResNet50V2 \citep{he2016DeepResidualLearning}                         & 0.89              & 0.90              \\
            MobileNet \citep{howard2017MobileNetsEfficientConvolutional}          & 0.91              & 0.91              \\
            MobileNetV2 \citep{sandler2018MobileNetV2InvertedResiduals}           & 0.87              & 0.88              \\
            EfficientNetB0 \citep{tan2019EfficientNetRethinkingModel}             & 0.20              & 0.19              \\
            MobileNetV3Large \citep{howard2019SearchingMobileNetV3}               & 0.58              & 0.53              \\
            MobileNetV3Small \citep{howard2019SearchingMobileNetV3}               & 0.31              & 0.32              \\
            RegNetX \citep{radosavovic2020DesigningNetworkDesign}                 & 0.18              & 0.12              \\
            EfficientNetV2B0 \citep{tan2021EfficientNetV2SmallerModels}           & 0.20              & 0.19              \\
            Mobile\_ViT \citep{mehta2022MobileViTLightweightGeneralpurpose}       & 0.55              & 0.54              \\
            ConvNeXtBase \citep{liu2022ConvNet2020s}                              & 0.78              & 0.76              \\ \hline
      \end{tabular}
\end{table}

\begin{figure}[htbp]
    \centering
    \begin{tabular}{cccc}
        \begin{subfigure}[b]{0.23\textwidth}
            \centering
            \includegraphics[width=\textwidth]{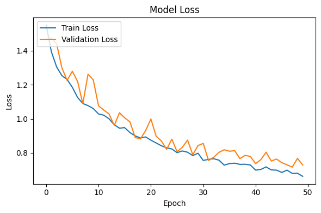}
            \caption{VGG16}
        \end{subfigure} &
        \begin{subfigure}[b]{0.23\textwidth}
            \centering
            \includegraphics[width=\textwidth]{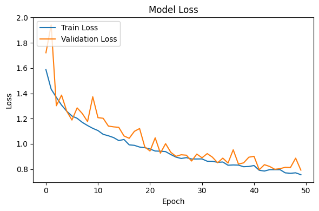}
            \caption{VGG19}
        \end{subfigure} &
        \begin{subfigure}[b]{0.23\textwidth}
            \centering
            \includegraphics[width=\textwidth]{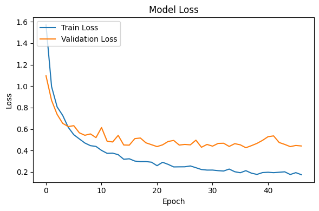}
            \caption{InceptionV3}
        \end{subfigure} &
        \begin{subfigure}[b]{0.23\textwidth}
            \centering
            \includegraphics[width=\textwidth]{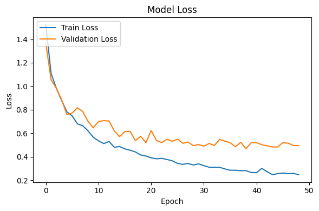}
            \caption{DenseNet121}
        \end{subfigure} \\
        
        \begin{subfigure}[b]{0.23\textwidth}
            \centering
            \includegraphics[width=\textwidth]{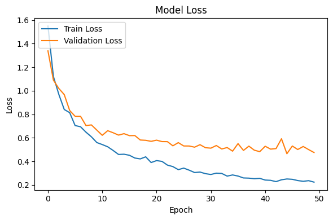}
            \caption{InceptionResNetV2}
        \end{subfigure} &
        \begin{subfigure}[b]{0.23\textwidth}
            \centering
            \includegraphics[width=\textwidth]{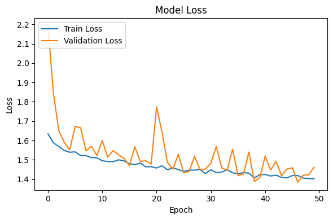}
            \caption{ResNet50}
        \end{subfigure} &
        \begin{subfigure}[b]{0.23\textwidth}
            \centering
            \includegraphics[width=\textwidth]{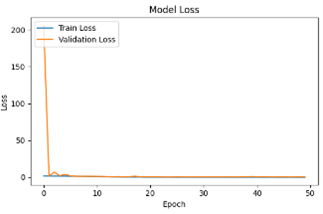}
            \caption{ResNet50V2}
        \end{subfigure} &
        \begin{subfigure}[b]{0.23\textwidth}
            \centering
            \includegraphics[width=\textwidth]{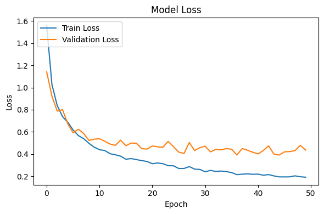}
            \caption{MobileNet}
        \end{subfigure} \\
        
        \begin{subfigure}[b]{0.23\textwidth}
            \centering
            \includegraphics[width=\textwidth]{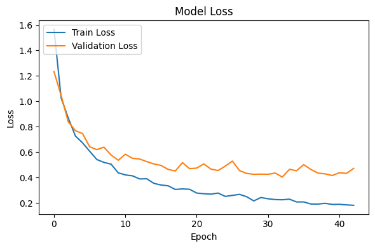}
            \caption{MobileNetV2}
        \end{subfigure} &
        \begin{subfigure}[b]{0.23\textwidth}
            \centering
            \includegraphics[width=\textwidth]{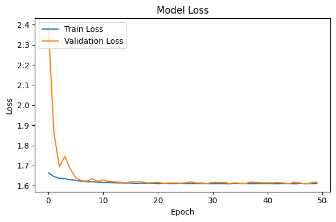}
            \caption{EfficientNetB0}
        \end{subfigure} &
        \begin{subfigure}[b]{0.23\textwidth}
            \centering
            \includegraphics[width=\textwidth]{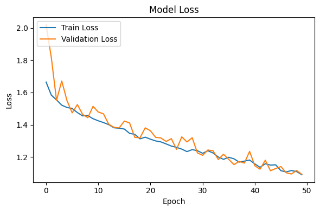}
            \caption{MobileNetV3Large}
        \end{subfigure} &
        \begin{subfigure}[b]{0.23\textwidth}
            \centering
            \includegraphics[width=\textwidth]{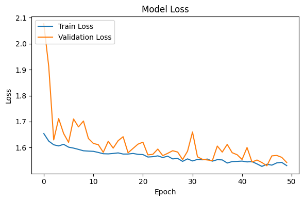}
            \caption{MobileNetSmall}
        \end{subfigure} \\
        
        \begin{subfigure}[b]{0.23\textwidth}
            \centering
            \includegraphics[width=\textwidth]{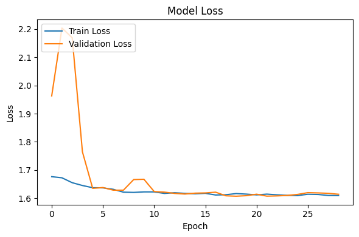}
            \caption{RegNetX}
        \end{subfigure} &
        \begin{subfigure}[b]{0.23\textwidth}
            \centering
            \includegraphics[width=\textwidth]{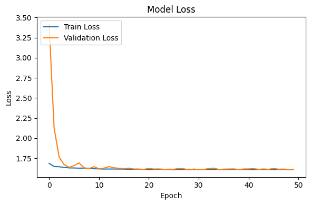}
            \caption{EfficientNetV2B0}
        \end{subfigure} &
        \begin{subfigure}[b]{0.23\textwidth}
            \centering
            \includegraphics[width=\textwidth]{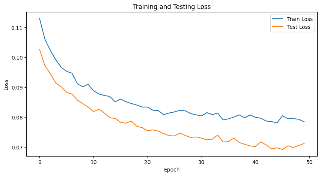}
            \caption{Mobile\_ViT}
        \end{subfigure} &
        \begin{subfigure}[b]{0.23\textwidth}
            \centering
            \includegraphics[width=\textwidth]{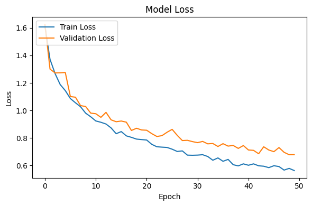}
            \caption{ConvNeXtBase}
        \end{subfigure}
    \end{tabular}
    \caption{Learning curve for different models}
    \label{fig:curves}
\end{figure}

To diagnose the training behavior of the models, learning curves (training loss + validation loss over epochs) for pavement extraction models are plotted in Figure~\ref{fig:curves}. Based on the observations of the learning curve for each model, it is evident that all models were trained well without showing signs of underfitting or overfitting during the training process. The learning curves depict the training loss and validation loss over 50 epochs for each model. For instance, the learning curves for models such as VGG16, VGG19, InceptionV3, DenseNet121, and InceptionResNetV2 show a consistent decrease in both training and validation losses, indicating stable learning and good generalization to the validation set. The MobileNet, MobileNetV2, and ResNet50V2 models exhibit particularly smooth curves with minimal fluctuation, suggesting robust training processes and high stability. These models reached convergence relatively quickly and maintained low validation losses, further supporting their strong performance metrics noted earlier. On the other hand, models like ResNet50 and EfficientNetB0 display higher variability in validation loss compared to training loss, which may indicate sensitivity to the validation set or potential overfitting that requires closer inspection. However, their overall trends still demonstrate effective learning without severe overfitting issues. MobileNetV3Large and MobileNetV3Small show some fluctuations in their validation loss curves, but they eventually stabilize, suggesting that with more epochs or fine-tuning, their performance could improve. RegNetX and EfficientNetV2B0 show high initial losses that decrease steadily, indicating that despite their initial high error rates, they do manage to learn effectively over time.

% \begin{figure}[htbp]
%       \centering
%       \includegraphics[width=1\linewidth]{image5.png} %todo: change this figure
%       \caption{Learning curve for different models}
%       \label{fig:curves}
% \end{figure}

After completing the comparison of the performance models, we then chose the top four models, ResNet50V2, InceptionV3, MobileNet, and DenseNet121, for the ensemble learning model. The ensemble model's performance is illustrated in the confusion matrix shown in Figure~\ref{fig:curves}. The accuracy of this ensemble model is 0.93, and the F1 score is 0.93.

The confusion matrix provides a detailed breakdown of the ensemble model's predictions across different categories: Fair, Good, Poor, Very Good, and Very Poor. The matrix reveals that the model achieved high accuracy across most categories. Specifically:

\begin{itemize}
      \item
            For the "Fair" category, the model correctly identified 160 instances without any misclassifications.
      \item
            In the "Good" category, the model correctly predicted 142 instances, with minimal misclassifications (3 instances classified as Fair and 6 as Very Good).
      \item
            For the "Poor" category, the model achieved perfect classification with all 147 instances correctly identified.
      \item
            The "Very Good" category had the most variance, with 121 correct predictions. There were some misclassifications: 12 instances were labeled as Fair, 24 as Good, 5 as Poor, and 2 as Very Poor.
      \item
            Lastly, for the "Very Poor" category, the model demonstrated perfect classification, correctly identifying all 153 instances.
\end{itemize}

Overall, the confusion matrix indicates that the ensemble model performs exceptionally well, with high precision and recall across all categories. This strong performance is reflected in the overall accuracy and F1 score of 0.93, confirming the effectiveness of combining the strengths of ResNet50V2, InceptionV3, MobileNet, and DenseNet121 in an ensemble approach. This ensemble model effectively balances the advantages of each individual model, leading to superior classification performance and robust predictions.

\begin{figure}[htbp]
      \centering
      \includegraphics[width=0.5\linewidth]{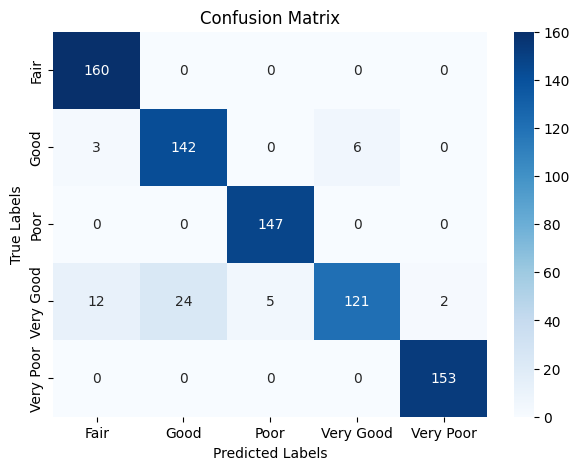}
      \caption{Confusion matrix of the ensemble model}
      \label{fig:matrix}
\end{figure}

\section{Conclusions}\label{conclusions}
This research focuses on the experimentation of satellite image analysis and its application in evaluating pavement conditions by exploring the deep learning based computer vision models in estimating road conditions at relatively large scales. These models have shown potential in identifying general pavement conditions. The use of 50 cm resolution satellite images in evaluating pavement conditions presents both challenges and opportunities. While this resolution is insufficient for detecting smaller pavement cracks directly, the aggregation of pixels can indicate the overall texture and pattern of the pavement surface, and deep learning models are capable of recognizing these patterns to infer the general condition of the pavement to highlight areas needing more detailed inspection, thus optimizing resource allocation for manual surveys. By considering the surrounding area and the overall appearance of the pavement, models can make broad predictions about its condition, which is useful for differentiating between condition states such as very good, good, fair, poor, and very poor. The case study results show that satellite-based infrastructure monitoring offers a promising and cost-effective approach for continuously monitoring pavement assets across extensive areas. By complementing traditional methods. This approach allows for extensive coverage without the need for on-site inspections, reducing both time and resource expenditure.

Although the initial application focused on a specific dataset, the framework is designed to be adaptable to various types of road conditions, including those influenced by unique climatic, geographic, or usage factors. Future work will involve training the models on a more diverse set of data, which includes different types of road environments, to enhance their robustness and accuracy across a broader spectrum of conditions.
It is important to note that this approach cannot replace vehicle-based pavement condition inspections, which provide much more detailed assessments, including specific measurements of cracks, rutting, and roughness. The current satellite image approach is only suitable for network-level quick estimates of the condition of the network, facilitating high-level decision-making. 
Future research could focus on enhancing satellite imagery analysis with additional data sources improves accuracy. Incorporating information such as traffic patterns, historical maintenance records, and climatic conditions can provide a comprehensive view of the pavement's health. This integrated approach enables more precise condition assessments.

% Bibliography
\bibliographystyle{unsrtnat}
\bibliography{references}

\end{document}